\documentclass[conference]{IEEEtran}
\IEEEoverridecommandlockouts
\usepackage{cite}
\usepackage{amsmath,amssymb,amsfonts}
\usepackage{algorithmic}
\usepackage{graphicx}
\usepackage{textcomp}
\usepackage{xcolor}
\usepackage{booktabs}
\usepackage{multirow} 
\usepackage{subcaption}
\usepackage{array}  
\usepackage{graphicx}
\def\BibTeX{{\rm B\kern-.05em{\sc i\kern-.025em b}\kern-.08em
    T\kern-.1667em\lower.7ex\hbox{E}\kern-.125emX}}

\usepackage[colorlinks=true, linkcolor=black, citecolor=black, urlcolor=black]{hyperref}

\begin{document}

\title{Inside CORE-KG: Evaluating Structured Prompting and Coreference Resolution for Knowledge Graphs
\thanks{All example names are fictitious and used solely to protect privacy.}
}

\author{\IEEEauthorblockN{Dipak Meher}
\IEEEauthorblockA{\textit{Department of Computer Science} \\
\textit{George Mason University}\\
Fairfax, USA \\
dmeher@gmu.edu}
\and
\IEEEauthorblockN{Carlotta Domeniconi}
\IEEEauthorblockA{\textit{Department of Computer Science} \\
\textit{George Mason University}\\
Fairfax, USA \\
cdomenic@gmu.edu}
}

\maketitle

\begin{abstract}
Human smuggling networks are increasingly adaptive and difficult to analyze. Legal case documents offer critical insights but are often unstructured, lexically dense, and filled with ambiguous or shifting references, which pose significant challenges for automated knowledge graph (KG) construction. While recent LLM-based approaches improve over static templates, they still generate noisy, fragmented graphs with duplicate nodes due to the absence of guided extraction and coreference resolution. The recently proposed CORE-KG framework addresses these limitations by integrating a type-aware coreference module and domain-guided structured prompts, significantly reducing node duplication and legal noise. In this work, we present a systematic ablation study of CORE-KG to quantify the individual contributions of its two key components. Our results show that removing coreference resolution results in a 28.25\% increase in node duplication and a 4.32\% increase in noisy nodes, while removing structured prompts leads to a 4.29\% increase in node duplication and a 73.33\% increase in noisy nodes. These findings offer empirical insights for designing robust LLM-based pipelines for extracting structured representations from complex legal texts.
\end{abstract}


\section{Introduction}
Human smuggling has become a complex operation involving dynamic networks of actors, routes, vehicles, and intermediaries \cite{carrasco2025scapegoating}. These networks facilitate illicit human mobility for payment, exploiting legal loopholes and adapting rapidly to enforcement efforts to avoid detection. Often linked to transnational crime, they pose serious security and humanitarian risks due to their covert, profit-driven nature. Understanding their structure is critical for shaping effective policy responses and disrupting illicit operations. However, much of the relevant intelligence remains locked within unstructured legal texts such as court rulings and case transcripts, making systematic analysis challenging.

Despite growing interest from legal and social science communities, computational approaches for analyzing such documents remain underdeveloped. Entity references in unstructured legal text are often inconsistent, appearing as varying coreferent mentions (e.g., “Defendant Lewis,” “smuggler,” “driver Lewis,” or simply “Lewis”), abbreviations, or role-based titles (e.g., “Officer Lewis” vs. “Defendant Lewis”). These inconsistencies complicate coreference resolution, entity normalization, and downstream tasks such as information extraction and knowledge graph construction.

Prior work has shown the value of knowledge graphs in legal investigations. Mazepa et al.~\cite{mazepa2022relationships} and Shi et al.~\cite{shi2022knowledge}, for instance, used rule-based methods to build graphs from legal cases, but their reliance on static templates limits flexibility in handling aliasing and surface-level entity variation. Large language models (LLMs) have shown promise in knowledge extraction~\cite{le2023large}, yet their use in human smuggling case narratives remains limited. A major challenge is merging semantically equivalent mentions of typed entities in complex legal texts~\cite{ji2020deep}, which, if left unresolved, creates fragmented and redundant graphs. Additionally, LLM-based extractors often misclassify or hallucinate entities and relations~\cite{zhang2024extract}, adding noise that hinders downstream analysis.

To address these challenges, we introduced CORE-KG~\cite{meher2025llm}, a modular framework that builds clean and interpretable knowledge graphs from legal case documents using large language models. CORE-KG combines two main components: (1) a type-aware coreference resolution module that sequentially resolves contextually similar mentions within each entity type, and (2) a structured prompting strategy that guides the model to extract relevant entities and relationships while filtering legal boilerplate and reducing ambiguity. Compared to a GraphRAG-based baseline, CORE-KG reduces node duplication by 33.28\% and legal noise by 38.37\%, showing strong improvements in graph quality.

In this paper, we conduct an ablation study on CORE-KG to quantify the individual contributions of its two key components: type-aware coreference resolution and structured prompting. In the first setting (\textit{CoreKG-no-coref}), we disable the coreference module while keeping structured prompts to assess its role in entity consolidation. In the second (\textit{CoreKG-no-structprompts}), we retain coreference resolution but remove structured prompts to evaluate their impact on reducing noise and improving extraction precision. We adopt the same setup as CORE-KG, using legal case documents related to human smuggling from publicly available U.S. federal and state court proceedings. Since no gold-standard graphs exist, we compare the outputs of each ablation with CORE-KG and a GraphRAG-based baseline~\cite{edge2024local}, both of which generate knowledge graphs from raw legal text using large language models.

Our ablation study reveals that removing either coreference resolution or structured prompts significantly degrades KG quality. Specifically, removing coreference resolution (\textit{CoreKG-no-coref}) results in a 28.25\% increase in node duplication and a 4.32\% increase in noisy nodes. In contrast, removing structured prompts (\textit{CoreKG-no-structprompts}) leads to a 73.33\% increase in noisy nodes and a 4.29\% increase in node duplication. On average, this corresponds to a 38.83\% increase in noise and a 16.27\% increase in node duplication, underscoring the necessity of both modules for producing accurate and interpretable knowledge graphs.\footnote{Our code is available at \url{https://github.com/dipakmeher/corekg-ablation-study.git.}}

\noindent To summarize, our key contributions are:
\begin{itemize}
    \item We present the first systematic ablation study of the CORE-KG framework, isolating the effects of its two central components: coreference resolution and structured prompting.
    
    \item We quantitatively show that removing the coreference module results in a 28.25\% increase in node duplication and 4.32\% increase in noise, while removing structured prompting leads to a 4.29\% increase in duplication and a 73.33\% spike in noise, demonstrating the distinct and complementary impact of each component.
    
    \item Our findings empirically validate that both components are essential and complementary for constructing accurate, interpretable knowledge graphs from unstructured legal texts, offering design guidance for future LLM-based KG pipelines.
\end{itemize}

\begin{figure}[t]
\centering
\includegraphics[width=\columnwidth]{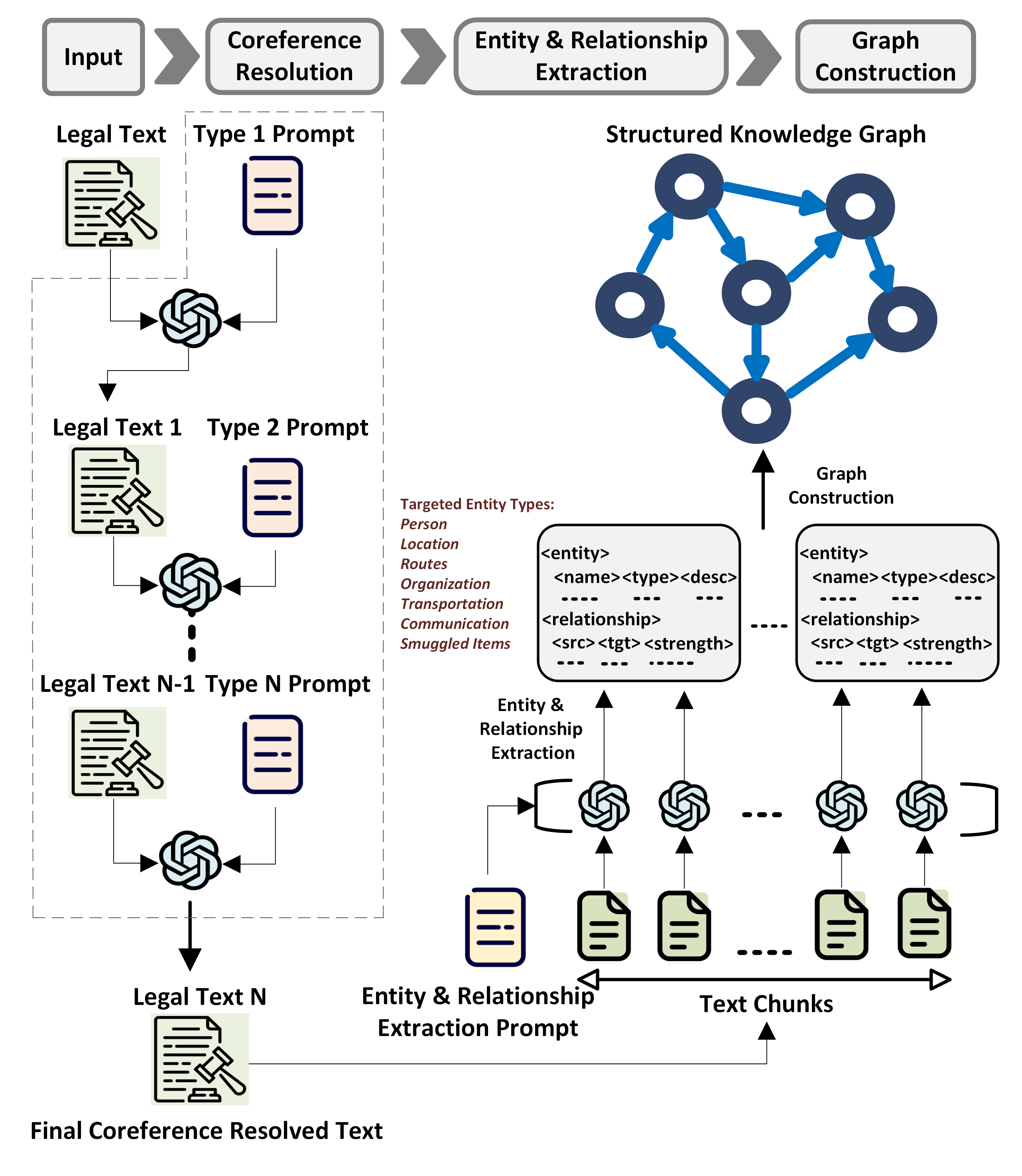}
\caption{Overview of the CORE-KG Pipeline. The pipeline begins with legal text inputs, processed via a type-aware coreference resolution module using sequential per-type prompting. The final resolved text is then passed through a structured prompting stage for entity and relationship extraction. The resulting triples are used to construct a coherent knowledge graph with significantly reduced duplication and noise.}
\label{fig:corekg_graph}
\end{figure}

\section{Related Work}

Knowledge graphs enable structured representation of unstructured text for tasks like reasoning, retrieval, and analysis~\cite{edge2024local}. They have seen wide adoption across domains including education~\cite{chen2018knowedu}, life sciences~\cite{callahan2024open}, and construction safety~\cite{fang2020knowledge}. However, poorly constructed graphs often exhibit node duplication and structural fragmentation, which hinder their effectiveness~\cite{huaman2020duplication}.

\subsection{Knowledge Graph Construction}

Traditionally, knowledge graphs are constructed by extracting key entities and their relationships from text and linking them sequentially. Entity extraction relies on rule-based, statistical, and domain-specific methods~\cite{sun2018overview}, while relation extraction uses syntactic, lexical, and semantic features~\cite{kambhatla2004combining}, ontology-based systems like HowNet~\cite{liu2007implementation}, and semi-supervised techniques~\cite{carlson2010coupled}. Though effective in narrow domains, these approaches often rely on hand-crafted rules and focus on maintaining close alignment with the source text.

More recently, LLM-based frameworks now enable prompt-driven knowledge graph construction with minimal manual effort~\cite{zhang2024extract, vizcarra2024representing}. Kommineni et al.~\cite{kommineni2024human} combined competency question generation, ontology design, and RAG-based triple extraction, while Zhang et al.~\cite{zhang2024extract} proposed a modular pipeline with open IE, schema definition, and canonicalization. However, these methods often assume clean input and overlook challenges like reference ambiguity and entity aliasing, which are common in legal texts where entities shift across aliases, roles, and pronouns. To address these, coreference resolution is critical for maintaining graph coherence. Wang et al.~\cite{wang2020coreference} demonstrated that even sparse pronoun usage can impact graph completeness, underscoring the need to integrate coreference mechanisms into the construction pipeline.

Several works have explored knowledge graph construction in criminal domains. Mazepa et al.~\cite{mazepa2022relationships} used rule-based NLP pipelines and CoreNLP to build a homicide investigation graph. Shi et al.~\cite{shi2022knowledge} employed regex-based extraction to construct a Neo4j graph for job-related crime indictments. While effective in legal contexts, both rely on static templates and lack coreference handling and modular prompting, limiting their robustness in complex narratives.

CORE-KG addresses this gap with a prompt-driven, entity-type-specific coreference resolution strategy tailored for modeling criminal networks. This design improves semantic precision and node coherence in knowledge graphs involving complex inter-entity relations such as migration routes and procedural actors. 

But to what extent do structured prompts and coreference resolution each contribute to graph quality in such complex domains? Our study aims to answer this through targeted ablations, isolating the individual and joint effects of these modules. By measuring changes in node duplication and legal noise, we empirically evaluate how each component supports accurate, interpretable knowledge graph construction.

\subsection{Coreference Resolution}
Several methods have been proposed to address node duplication and fragmentation via coreference resolution~\cite{liu2023brief}. CNN-based models by Pogorilyy et al.~\cite{pogorilyy2019coreference} and Wu et al.~\cite{wu2017deep} enhance semantic and syntactic pattern modeling.

Recent advances in large language models (LLMs) have advanced coreference resolution in low-resource settings, using prompt-based, few/zero-shot, and chain-of-thought strategies~\cite{das2024co, gan2024assessing}. However, these methods often target single entity types or general narratives without modeling type distinctions~\cite{das2024co, tran2025coreference}. While effective in focused contexts, their ability to generalize across diverse, role-shifting entity types in domain-specific texts remains underexplored.

Some progress has been made in coreference resolution for legal texts. Jia et al.~\cite{ji2020deep} introduced a neural model combining ELMo, BiLSTM, and GCNs to resolve speaker-based coreference in court records. However, to the best of our knowledge, no prior work has examined coreference resolution in the legal domain using large language models. While instruction-tuned LLMs have shown competitive performance in coreference resolution~\cite{le2023large}, these studies are restricted to standard narrative data and do not address the challenges of multi-entity legal documents. This raises an open question: to what extent can LLM-based coreference resolution scale to complex, multi-entity legal texts, and how much does it contribute to improving knowledge graph quality?

\section{Method}

\subsection{CORE-KG Overview}

Figure~\ref{fig:corekg_graph} illustrates the complete CORE-KG pipeline. CORE-KG consists of two key components: (1) a \textit{type-aware coreference resolution} module that consolidates semantically and contextually equivalent mentions within each entity type, and (2) a \textit{knowledge graph construction} module that employs structured prompts to extract entities and relationships. These prompts incorporate domain-specific filtering instructions to suppress legal boilerplate, sequential type-wise extraction to minimize attention drift, and explicit type definitions to reduce classification ambiguity.

\subsubsection{Coreference Resolution}

We design a Coreference Resolution module that leverages the contextual reasoning capabilities of large language models (LLMs)~\cite{wang2023can,havrilla2024glore}. The model unifies disparate mentions under a canonical form (e.g., “Young” and “the defendant”), ensuring that all relational links converge on a single entity node. This improves structural consistency and enhances analytical clarity.

However, resolving all entity types simultaneously can dilute the model’s attention and cause type drift (incorrect category assignments) or feature entanglement (interference between features of different types)~\cite{zhou2023universalner, abdelnabi2024you}. For instance, the phrase “The Camp” might denote a temporary migrant holding site (\textit{Location}) but may be misclassified as an \textit{Organization} in ambiguous contexts. Similarly, “the van” might refer to either a transport vehicle (\textit{Means of Transportation}) or a holding site (\textit{Location}).

To mitigate these issues, we adopt a \textit{type-wise sequential resolution strategy}, where the LLM resolves one entity type at a time. This reduces cross-type interference and allows more accurate coreference consolidation. The pipeline proceeds as follows:
\begin{enumerate}
    \item The legal text is first passed to the LLM with a prompt focused type Person, linking mentions like ``Young,'' ``the defendant,'' and ``the driver'' to a canonical form.
    \item The output is then processed with a Location prompt to resolve place-based mentions (e.g., ``Laredo,'' ``Laredo Texas'').
    \item This sequence continues for Routes, Organization, Means of Transportation, Means of Communication, and Smuggled Items.
\end{enumerate}

This modular strategy enhances \textit{semantic precision} and \textit{structural fidelity} while aligning with criminal network modeling through type-limited resolution guided by domain expertise.

\paragraph{Coreference Resolution Prompt Design}
\label{sec:coref_resolution_prompt_design}

Prompts are designed to generalize across diverse legal cases while capturing the variability in how entities appear in text. For instance, Person entities may include aliases, shortened names, or role-based mentions (e.g., “the driver”); Location entities may vary in granularity (e.g., “Laredo” vs. “Laredo, Texas”); Routes may be expressed using formal names or abbreviations (e.g., “Interstate 35” vs. “I-35”); and Means of Transportation often combine object references with ownership cues (e.g., “trailer” vs. “the defendant’s truck”).

To ensure consistency and precision, each prompt follows a structured format that includes a persona definition, clear task description, contextual information, entity-type-specific resolution rules, and few-shot examples. These components guide the LLM to resolve coreferences accurately without altering the input while accommodating the different ways each entity type is expressed in legal documents.

\subsubsection{Entity-Relationship Extraction and KG Construction}

The coreference-resolved text is fed into the Knowledge Graph Construction (KGC) module of the GraphRAG framework \cite{edge2024local}, which comprises: (1) a KGC component that extracts entities and relationships to form a structured graph, and (2) a retrieval module that uses this graph as an index for response generation. This work focuses on the KGC component, extended for processing legal case files on human smuggling.

Each document is split into overlapping 300-token chunks, passed to the LLM with a domain-optimized extraction prompt. The model returns entity–relationship triples, which are aggregated across chunks. Post-processing merges entities by exact string and type match, and the graph is built using \texttt{NetworkX}. Outputs are saved in GraphML and Parquet formats for downstream analysis in tools like Gephi and the GraphRAG visualizer.

\paragraph{Prompt Tuning for Entity and Relationship Extraction} 
\label{sec:prompt_tuning}

The extraction prompt follows the same structured design as described in Section~\ref{sec:coref_resolution_prompt_design}, but is adapted for a single-pass setup required by GraphRAG’s KGC component. Unlike coreference resolution, which uses separate prompts per entity type, this unified prompt extracts all seven entity types and their relationships together. It includes the same key components, persona, task description, contextual framing, step-by-step instructions, output format, and few-shot examples, to ensure relevance and reduce noise.

Additional refinements are introduced to improve precision for human smuggling network analysis, as outlined below.

\textit{(1) Sequential Entity Extraction to Reduce Attention Spread:}  
GraphRAG’s default joint extraction can dilute model attention across entity types, leading to missed entities, misclassifications, or loss of fine-grained distinctions, issues common in legal texts with overlapping entity mentions \cite{abdelnabi2025get}. To address this, we enforce a fixed extraction order: \textit{Person} first, followed by \textit{Location}, \textit{Routes}, \textit{Organization}, etc. The model extracts relationships only after completing all entity types in sequence. This reduces attention competition, improves entity precision, and ensures more complete and reliable knowledge graph construction.

\textit{(2) Filtering High-Frequency Irrelevant Entities: }  
Legal texts often include frequently mentioned but non-critical entities, such as courts, juries, and appeals, typically extracted as \textit{Organization} type. If retained, these inflate the graph with irrelevant nodes and distort relationship patterns. To prevent this, we add an explicit filtering step in the prompt. After extraction, the LLM removes all government-related entities based on predefined rules before generating the final output. This in-prompt filtering reduces noise, improves graph clarity, and eliminates the need for post-processing.

\textit{(3) Entity Type Definitions to Reduce Misclassification: }  
LLMs often misclassify entities due to overgeneralization e.g., labeling event names as \textit{Location} just because they appear near geographic terms during training \cite{peters2025generalization, dai2024bias}. This problem worsens in legal texts, where subtle context matters. To reduce such errors, we include clear definitions and examples for all seven entity types directly in the prompt. These definitions help the model distinguish between similar types like \textit{Person}, \textit{Organization}, and \textit{Route}, leading to more accurate and context-aware extractions.

\subsection{Ablation Study: Isolating the Impact of Coreference Resolution and Structured Prompts}

To evaluate the individual contributions of the core components of CORE-KG, we design a focused ablation study that isolates the effects of (1) type-aware coreference resolution and (2) domain-specific structured prompts. While the full CORE-KG pipeline incorporates both modules, this study aims to disentangle their respective influence on the quality of the resulting knowledge graphs and assess how each module contributes to cleaner, more complete, and more accurate representations of human smuggling networks.

We construct three experimental variants for comparison. The first is the Full CORE-KG pipeline, which includes both coreference resolution and structured prompts. The second variant, denoted as CoreKG-no-coref, removes the coreference module while retaining the structured prompt for entity and relationship extraction. In this setting, raw legal text is passed directly to the KGC module without resolving intra-document entity references. The third variant, CoreKG-no-structprompt, removes our structured extraction prompt and instead uses the default prompt from GraphRAG \cite{edge2024local}, while still applying coreference resolution. This allows us to isolate the contribution of structured prompt design while holding the input (coref-resolved text) constant.

Finally, we include the GraphRAG baseline, which uses neither coreference resolution nor structured prompts and instead relies on its default entity and relationship extraction settings, with only minimal adaptation through a few in-domain examples from human smuggling cases. Including GraphRAG allows us to anchor our ablation study to a standard LLM-based extraction pipeline. By comparing CoreKG-no-coref and CoreKG-no-structprompt against this baseline, we can directly quantify how each individual module (coreference resolution or structured prompting) improves graph quality over a non-specialized system, and how the full CORE-KG pipeline combines these gains to achieve the strongest performance.

\begin{table*}[t]
\centering
\setlength{\tabcolsep}{3pt}
\fontsize{6.4pt}{8.9pt}\selectfont 
\begin{tabular}{l|
ccc|ccc|ccc|ccc||
ccc|ccc|ccc|ccc}
\multirow{2}{*}{Case} 
& \multicolumn{12}{c||}{\textbf{Node Duplication}} 
& \multicolumn{12}{c}{\textbf{Noisy Nodes}} \\
\cmidrule(lr){2-13} \cmidrule(lr){14-25}
& \multicolumn{3}{c|}{GraphRAG} & \multicolumn{3}{c|}{CoreKG-no-coref} & \multicolumn{3}{c|}{CoreKG-no-str-prompt} & \multicolumn{3}{c||}{CoreKG}
& \multicolumn{3}{c|}{GraphRAG} & \multicolumn{3}{c|}{CoreKG-no-coref} & \multicolumn{3}{c|}{CorekG-no-str-prompt} & \multicolumn{3}{c}{CoreKG} \\
\cmidrule(lr){2-4} \cmidrule(lr){5-7} \cmidrule(lr){8-10} \cmidrule(lr){11-13}
\cmidrule(lr){14-16} \cmidrule(lr){17-19} \cmidrule(lr){20-22} \cmidrule(lr){23-25}
& Tot & Dup & Rate & Tot & Dup & Rate & Tot & Dup & Rate & Tot & Dup & Rate
& Tot & Noisy & Rate & Tot & Noisy & Rate & Tot & Noisy & Rate & Tot & Noisy & Rate \\
\midrule
Case 1  & 94 & 32 & 34.04 & 63 & 20 & 31.75 & 75 & 21 & 28 & 51 & 13 & \textbf{25.49}  & 94 & 26 & 27.66 & 63 & 8 & 12.70 & 75 & 13 & 17.33 & 51 & 5 & \textbf{9.80} \\
Case 2  & 86 & 24 & 27.91 & 48 & 11 & 22.92 & 81 & 15 & 18.52 & 41 & 5 & \textbf{11.9}  & 86 & 28 & 32.56 & 48 & 2 & 4.17  & 81 & 30 & 37.04 & 42 & 0 & \textbf{0.00} \\
Case 3  & 68 & 19 & 27.94 & 37 & 11 & 29.73 & 63 & 13 & 20.63 & 32 & 6 & \textbf{18.75} & 68 & 38 & 55.88 & 37 & 9 & \textbf{24.32} & 63 & 36 & 57.14 & 32 & 13 & 40.63 \\
Case 4  & 99 & 30 & 30.3 & 71 & 24 & 33.8 & 84 & 26 & 30.95 & 63 & 19 & \textbf{30.16}  & 99 & 12 & 12.12 & 71 & 3 & \textbf{4.23}  & 84 & 13 & 15.48 & 63 & 3 & 4.76 \\
Case 5  & 66 & 21 & 31.82  & 50 & 17 & 34 & 86 & 20 & \textbf{23.26} & 60 & 19 & 31.67  & 66 & 15 & 22.73 & 50 & 8 & \textbf{16.00} & 86 & 16 & 18.60 & 60 & 11 & 18.33 \\
Case 6  & 83 & 33 & 39.76 & 59 & 19 & 32.2 & 76 & 12 & 15.79 & 32 & 4 & \textbf{6.25}  & 83 & 9 & 10.84 & 59 & 9 & 15.25 & 76 & 12 & 15.79 & 32 & 4 & \textbf{6.25} \\
Case 7  & 87 & 33 & 37.93  & 67 & 24 & 35.82 & 79 & 17 & 21.52 & 49 & 10 & \textbf{20.41}  & 87 & 12 & 13.79 & 67 & 7 & \textbf{10.45} & 79 & 15 & 18.99 & 49 & 8 & 16.33 \\
Case 8  & 60 & 19 & 31.67 & 29 & 7 & 24.14 & 40 & 10 & 25 & 22 & 5 & \textbf{22.73}  & 60 & 17 & 28.33 & 29 & 11 & 37.93 & 40 & 11 & 27.50 & 22 & 5 & \textbf{22.73} \\
Case 9  & 75 & 15 & 20 & 22 & 1 & \textbf{4.55} & 61 & 9 & 14.75 & 34 & 5 & 14.71 & 75 & 19 & 25.33 & 22 & 3 & \textbf{13.64} & 61 & 18 & 29.51 & 34 & 7 & 20.59 \\
Case 10 & 76 & 18 & 23.68 & 29 & 7 & 24.14 & 65 & 12 & 18.46 & 23 & 4 & \textbf{17.39} & 76 & 40 & 52.63 & 29 & 14 & 48.28 & 65 & 45 & 69.23 & 23 & 10 & \textbf{43.48} \\
Case 11 & 49 & 11 & 22.45 & 28 & 5 & 17.86 & 30 & 6 & \textbf{13} & 21 & 3 & 14.29  & 49 & 9 & 18.37  & 28 & 5 & 17.86 & 30 & 5 & 16.67 & 21 & 3 & \textbf{14.29} \\
Case 12 & 68 & 20 & 29.41  & 46 & 8 & \textbf{17.39} & 63 & 13 & 20.63 & 57 & 15 & 26.32  & 68 & 10 & 14.71 & 46 & 4 & \textbf{8.70}  & 63 & 9 & 14.29 & 57 & 5 & 8.77 \\
Case 13 & 103 & 29 & 28.16 & 47 & 11 & 23.4 & 91 & 14 & 15.38 & 41 & 5 & \textbf{12.2} & 103 & 49 & 47.57 & 47 & 18 & 38.30 & 91 & 40 & 43.96 & 41 & 11 & \textbf{26.83} \\
Case 14 & 55 & 13 & 23.64 & 36 & 7 & 19.44 & 37 & 4 & 10.81 & 28 & 1 & \textbf{3.57}  & 55 & 10 & 18.18 & 36 & 4 & 11.11 & 37 & 5 & 13.51 & 28 & 3 & \textbf{10.71} \\
Case 15 & 73 & 26 & 35.62   & 57 & 19 & 33.33 & 72 & 16 & \textbf{22.22} & 49 & 11 & 22.45  & 73 & 10 & 13.70 & 57 & 5 & \textbf{8.77}  & 72 & 11 & 15.28 & 49 & 5 & 10.20 \\
Case 16 & 77 & 26 & 33.77  & 42 & 10 & \textbf{23.81} & 95 & 27 & 28.42 & 51 & 15 & 29.41  & 77 & 17 & 22.08 & 42 & 6 & \textbf{4.29} & 95 & 32 & 33.68 & 51 & 7 & 13.73 \\
Case 17 & 78 & 17 & 21.79   & 56 & 11 & 19.64 & 70 & 9 & 12.86 & 47 & 6 & \textbf{12.77}  & 78 & 10 & 12.82 & 56 & 4 & 7.14  & 70 & 8 & 11.43 & 47 & 3 & \textbf{6.38} \\
Case 18 & 71 & 19 & 26.76 & 45 & 9 & 20 & 57 & 11 & 19.3 & 33 & 6 & \textbf{18.18} & 71 & 25 & 35.21 & 45 & 11 & 24.44 & 57 & 24 & 42.11 & 33 & 5 & \textbf{15.15} \\
Case 19 & 94 & 36 & 38.3 & 29 & 8 & 27.59 & 83 & 17 & \textbf{20.48} & 41 & 9 & 21.95 & 94 & 49 & 52.13 & 29 & 5 & \textbf{17.24} & 83 & 43 & 51.81 & 41 & 8 & 19.51 \\
Case 20 & 81 & 37 & 45.68 & 56 & 25 & \textbf{44.64} & 61 & 22 & 36.07 & 49 & 22 & 44.9  & 81 & 26 & 32.10 & 56 & 7 & \textbf{12.50} & 61 & 17 & 27.87 & 49 & 9 & 18.37 \\
\midrule
\textbf{Avg} & 77.15 & 23.9 & 30.53 & 45.85 & 12.4 & 26.01 & 68.45 & 14.7 & 21.15 & 41.25 & 9.05 & \textbf{20.27} & 77.15 & 21.55 & 27.43 & 45.85 & 7.15 & 17.37 & 68.45 & 20.15 & 28.86 & 41.25 & 6.25 & \textbf{16.65} \\
\end{tabular}
\caption{
Comparison of node duplication and legal noise across different extraction methods. The table includes two ablation settings derived from CORE-KG, and baselines (CORE-KG and GraphRAG). 
\textit{Tot} indicates the total number of extracted entities; \textit{Dup} denotes the number of duplicate or noisy entities; \textit{Rate} represents the percentage of duplication or noise relative to the total. 
Bold values indicate the lowest Rate within each group of methods.
}
\label{tab:results-duplicate-noise}
\end{table*}

\section{Experimental Setup}

We aim to answer the following research questions:
\begin{itemize}
   \item RQ1: How much does the coreference resolution module contribute to reducing duplicate nodes and noisy entities in knowledge graphs constructed from human smuggling cases?  

   \item RQ2: How much does the modified structured prompt contribute to reducing duplicate nodes and noisy entities in knowledge graphs?  
\end{itemize}

\subsection{Dataset}

We follow the same dataset setup as the CORE-KG framework to ensure a consistent and fair comparison across system variants. The dataset consists of judicial case documents related to human smuggling, retrieved from the Nexis Uni academic search engine. These cases span U.S. federal and state court proceedings filed between 1994 and 2024, covering a diverse range of smuggling-related scenarios.

Since legal case files often contain lengthy procedural and statutory sections, we extract only the ``Opinion'' section from each case, which typically includes the most relevant factual narrative, such as involved individuals, routes, vehicles, and smuggled items. This section serves as the input for all graph construction experiments including GraphRAG (used as a baseline in CORE-KG), the ablation variants, and the full CORE-KG pipeline.

For this study, we use the same 20 legal cases as in CORE-KG. Each case contains approximately 2000 words in the Opinion section and reflects the narrative complexity necessary for evaluating the impact of coreference resolution and structured prompting on knowledge graph quality.

\subsection{Implementation}

We use the LLaMA 3.3 70B model for both coreference resolution and knowledge graph construction, served locally via the Ollama framework with temperature set to zero to ensure deterministic outputs. All experiments are conducted on an NVIDIA A100 GPU with 80GB of memory.

The knowledge graph construction is based on the GraphRAG framework (v0.3.2), configured with overlapping 300-token chunks. Although the framework requires specifying an embedding model, such as \texttt{nomic-embed-text} in our case, we do not use embeddings in this work, as the graph construction process does not rely on them. The codebase is implemented in Python 3.12.

\subsection{Baselines and Experimental Design}

To evaluate the individual contributions of the two core modules in the CORE-KG pipeline, coreference resolution and structured prompting, we conduct an ablation study comparing four system variants:

\begin{itemize}
    \item GraphRAG: A minimally adapted version of the GraphRAG framework \cite{edge2024local}, using its standard prompt. It is applied directly to the input text, without coreference resolution or modified structured prompts.

    \item CoreKG-no-coref: Applies the structured prompt from CORE-KG to the raw input text, without any coreference resolution. This variant isolates the impact of the modified structured prompt on knowledge graph construction.

    \item CoreKG-no-str-prompt: Applies the type-aware coreference resolution module but uses the standard GraphRAG prompt for entity and relationship extraction. This variant isolates the impact of coreference resolution alone.

    \item CORE-KG (Full): The complete pipeline with both modules enabled—type-aware coreference resolution followed by entity and relationship extraction guided by the modified structured prompt. This represents the end-to-end system evaluated in the original CORE-KG framework.
\end{itemize}

All systems use the same input documents, model (LLaMA 3.3 70B), and GraphRAG-based graph construction setup to ensure a fair comparison.

\subsection{Evaluation Measures}

Since no gold-standard annotations exist for knowledge graphs in this domain, we adopt a quantitative evaluation framework using two metrics: node duplication rate and noise rate. These metrics are computed for all system variants using the same 20 legal cases from the CORE-KG setup.

Node duplication rate measures the proportion of redundant entity nodes in the graph. Duplicate detection is performed in two stages: (1) intra-type fuzzy string matching using the \texttt{partial\_ratio} function from the \texttt{RapidFuzz} library, with a similarity threshold of 75\%; and (2) manual review by a domain expert to refine the clusters. Given a set of duplicate clusters $C_i$, the duplication count is computed as $\sum_{C_i} (|C_i| - 1)$, and normalized by total node count.

Noise rate quantifies the percentage of extracted nodes that are irrelevant to smuggling network analysis, such as legal boilerplate terms (e.g., Court, Appeal Process, Judicial Proceedings). These are manually identified by a domain expert and reported as a proportion of total nodes.

In addition to these metrics, we perform a graph-level comparison across CORE-KG, CoreKG-no-coref, and CoreKG-no-str-prompt. This analysis provides qualitative insight into the nature of duplication, noise, relational cohesion, and semantic coherence, offering a deeper understanding of how different components impact overall graph quality.

\section{Results}
Table~\ref{tab:results-duplicate-noise} reports node duplication and noisy nodes statistics across 20 legal case graphs, comparing the outputs of four configurations: GraphRAG, CoreKG-no-coref (structured prompts without coreference resolution), CoreKG-no-str-prompt (coreference resolution with the standard GraphRAG prompt), and the full CoreKG pipeline.

\subsection{RQ1: Impact of Coreference Resolution on Node Duplication and Noise}
\label{sec:rq1_CoreKG-no-coref}
\subsubsection{Node Duplication}

As shown in Table~\ref{tab:degradation}, removing the coreference resolution module leads to a notable degradation due to increased duplicate nodes. CoreKG-no-coref exhibits a duplication rate of 26.01\%, compared to 20.28\% in CORE-KG, reflecting a relative degradation of 28.25\%. The GraphRAG baseline shows the highest duplication at 30.53\%, underscoring the effectiveness of both coreference resolution and structured prompting in minimizing redundancy.

As shown in Table~\ref{tab:results-duplicate-noise}, case-wise analysis reinforces this trend. In nearly all cases, CoreKG-no-coref exhibits higher duplication than CORE-KG. For example, in Case~6, the duplication rate rises from 6.25\% (CORE-KG) to 35.82\%, and in Case~13, from 3.57\% to 19.44\%. The only exceptions are Case~9, Case~12, Case~16, and Case~20, where CoreKG-no-coref reports slightly lower rates. However, these differences stem from minor fluctuations in the total number of nodes extracted by the LLM, which in turn reduce the observed duplication proportion. Overall, on average, CORE-KG maintains lower duplication rates compared to the ablation variants and GraphRAG.




\begin{table}[h]
\centering
\caption{Average node duplication and noise rates along with relative degradation compared to CORE-KG, computed over 20 legal cases}
\label{tab:degradation}
\begin{tabular}{lcc}
\toprule
\textbf{Method} & \textbf{Node Duplication (\%)} & \textbf{Noisy Nodes (\%)} \\
\midrule
GraphRAG                & 30.53\% (+50.54\%) & 27.43\% (+64.74\%) \\
CoreKG-no-coref         & 26.01\% (+28.25\%) & 17.37\% (+4.32\%) \\
CoreKG-no-structprompts & 21.15\% (+4.29\%)  & 28.86\% (+73.33\%) \\
\textbf{CORE-KG} & \textbf{20.28\%} (–) & \textbf{16.65\%} (–) \\
\bottomrule
\end{tabular}
\end{table}

In CoreKG-no-coref, using structured prompts alone already reduces duplication compared to GraphRAG, lowering the average node-duplication rate from 30.53\% to 26.01\%, as shown in Table~\ref{tab:degradation}. This reduction comes primarily from the in-prompt filtering rules, sequential extraction steps, and the explicit definition of entity types relevant to human smuggling analysis.

However, structured prompts alone cannot resolve referential variations across the text. As shown in the CoreKG-no-coref graph for Case~6 in Figure~\ref{fig:abl1_graph}, the mentions “Interstate Highway 35,” “Interstate,” and “I-35” all refer to the same route. Without coreference resolution, these surface variations are extracted as separate nodes, fragmenting what should be a single entity. This leads to a higher node-duplication rate in CoreKG-no-coref (26.01\%) compared to the full CORE-KG pipeline (20.28\%).

\subsubsection{Noisy Nodes}

The removal of coreference resolution leads to a moderate increase in noisy nodes, rising from 16.65\% in CORE-KG to 17.37\% in CoreKG-no-coref, a relative degradation of 4.32\% (Table~\ref{tab:degradation}). This shows that coreference resolution helps reduce noise, though its effect is smaller than on duplication. It refines the input text by consolidating scattered or semantically equivalent mentions that might otherwise increase text density, diffuse the LLM’s attention, and result in noisier extractions or misclassifications. While the improvement is modest, it supports a clearer and more accurate graph.

Even without coreference resolution, CoreKG-no-coref maintains comparatively low noise levels. This indicates that the structured prompt plays a key role in reducing noise by guiding the LLM to extract relevant and precise entities and relationships. Through in-prompt filtering, sequential extraction, and clearly defined entity-type categories, the structured prompt suppresses irrelevant or overly generic mentions. The sharp rise in noise from CoreKG-no-coref (17.37\%) to GraphRAG (27.43\%) highlights the importance of structured prompting, since GraphRAG does not include these additional prompt-level instructions, resulting in poorer performance.

In 5 out of 20 cases, CoreKG-no-coref achieves noise rates comparable to or even lower than CORE-KG, suggesting that structured prompts alone can sometimes perform competitively. Nonetheless, the full CORE-KG pipeline consistently yields the lowest noise levels, reaffirming that both components are complementary and jointly essential for high-precision knowledge graph construction.

\subsection{RQ2: Impact of CORE-KG's Structured Prompt on Node Duplication and Noise}
\label{subsec:rq2corekgnostrprompt}
\subsubsection{Node Duplication}

As shown in Table~\ref{tab:degradation}, removing the structured prompt in CoreKG-no-str-prompt increases the node duplication rate from 20.28\% (CORE-KG) to 21.15\%, a relative degradation of approximately 4.29\%. This highlights the role of CORE-KG’s structured prompt in guiding the LLM toward precise, domain-relevant entity extraction and in mitigating node duplication, even when coreference resolution is still present.

The structured prompt in CORE-KG employs sequential extraction and explicit type definitions to mitigate attention drift. This design prevents the LLM from attending to multiple entity and relationship types simultaneously, which can lead to over-extraction or conflated outputs. Without such structured guidance, as in CoreKG-no-str-prompt, the model may extract longer or imprecise entities that inflate the number of nodes. For example, in the sentence “The driver and the smuggler entered Texas with the group,” the model may extract relevant entities such as “the group” and “Texas,” but also over-extract the phrase “entered Texas with the group” as a separate entity. Although this phrase refers to the same underlying entities, its inclusion introduces duplication due to loss of precision in entity extraction, driven by attention drift.

\begin{figure*}[htbp]
    \centering
    \begin{subfigure}[t]{0.4008\textwidth}
        \hspace{-1.2cm}
        \includegraphics[width=1.4\linewidth]{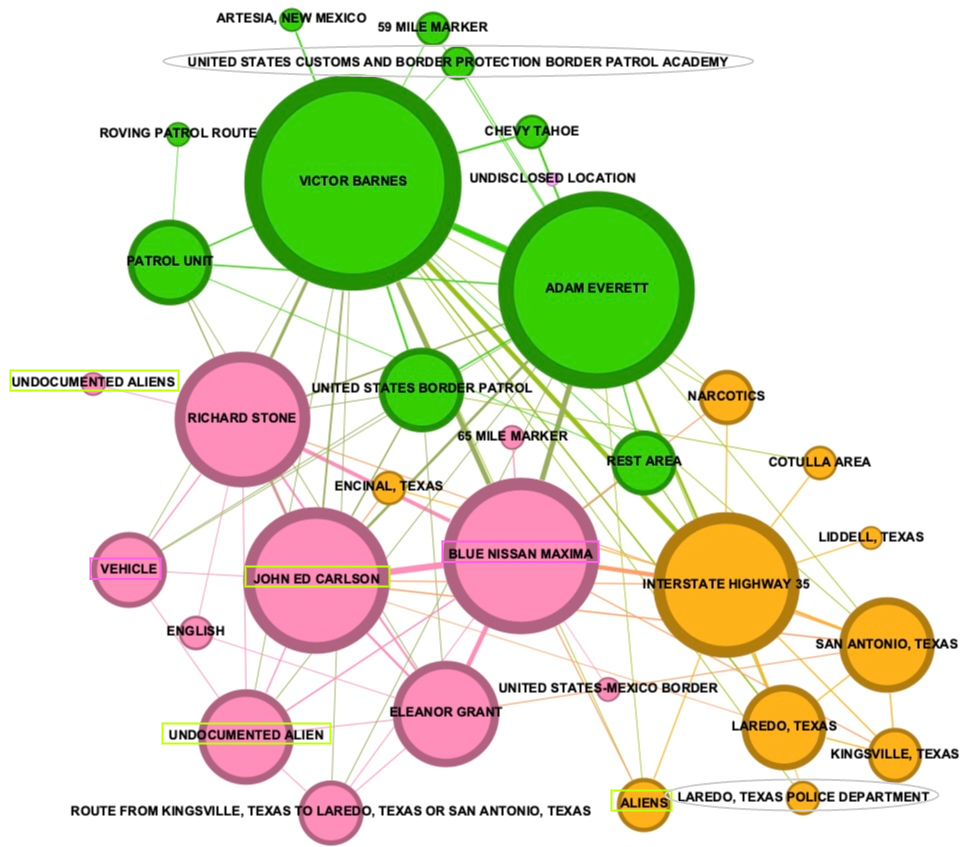}
        \caption{CORE-KG Graph}
        \label{fig:corekg_graph}
    \end{subfigure}
    \hfill
    \begin{subfigure}[t]{0.5\textwidth}
        \centering
        \includegraphics[width=1.18\linewidth]{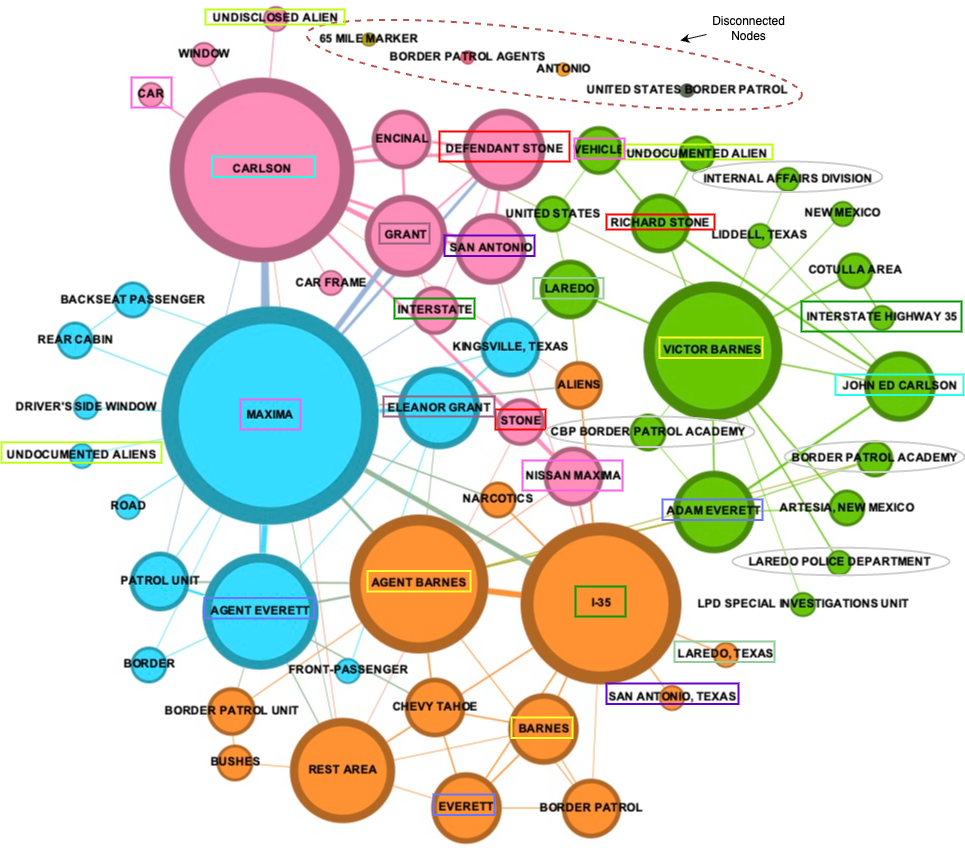}
        \caption{CoreKG-no-coref Graph}
        \label{fig:abl1_graph}
    \end{subfigure}
    \hfill
    \begin{subfigure}[t]{0.90\textwidth}
        \vspace{0.5cm}
        \includegraphics[width=\linewidth]{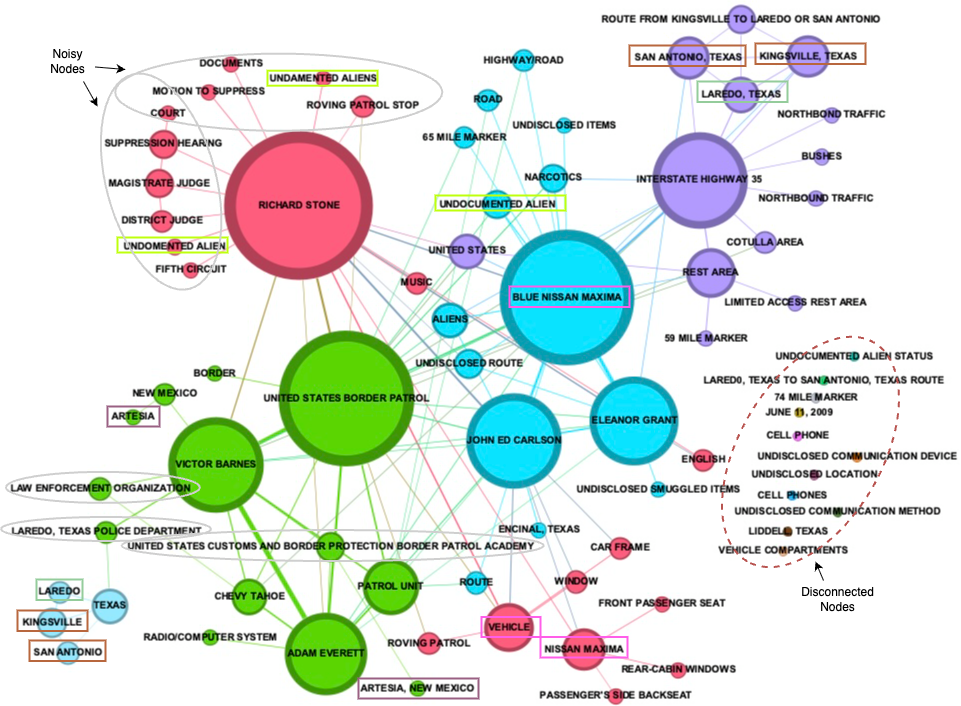}
        \caption{CoreKG-no-str-prompt Graph}
        \label{fig:abl2_graph}
    \end{subfigure}
    \caption{Graphs of (a) CORE-KG; (b) CoreKG-no-coref; (c) CoreKG-no-struct-prompt, shown for Case 06. Duplicate entities are indicated with solid rectangles, noisy or irrelevant entities with solid ovals, and disconnected or weakly connected nodes with dashed ovals.}
    \label{fig:graph_comparison}
\end{figure*}

Case-wise analysis (Figure~\ref{fig:graph_comparison}) further supports this conclusion. CoreKG-no-str-prompt shows higher duplication than CORE-KG in most cases. Notably, Case~06 demonstrates the largest increase, with node duplication rising from 6.25\% in CORE-KG to 15.79\% in CoreKG-no-str-prompt. This suggests that structured prompting and coreference resolution are jointly necessary for achieving substantial reductions in duplication.

However, there are six cases (Case~06, Case~12, Case~15, Case~16, Case~19, and Case~20) where CoreKG-no-str-prompt shows lower duplication than CORE-KG. This can be explained by two factors. First, even without structured prompting, the coreference module still mitigates some duplication by merging exact or near-exact mentions. Second, the absence of structured prompts leads to a surge in extracted nodes, many of which are noisy or semantically conflated. This broader and less focused extraction dilutes the proportion of detected duplicates, giving the appearance of a lower duplication rate despite the underlying graph being noisier. As shown in Table~\ref{tab:results-duplicate-noise}, the average number of extracted nodes increases from 41.25 in CORE-KG to 68.45 in CoreKG-no-str-prompt, contributing to the observed variations in duplication percentages.

\subsubsection{Noisy Nodes}

CoreKG-no-str-prompt exhibits a substantial increase in the proportion of noisy nodes, rising from 16.65\% in CORE-KG to 28.86\%, a relative degradation of 73.33\%, as shown in Table~\ref{tab:degradation}. This spike is primarily due to the absence of structured prompting, which normally provides in-prompt filtering, sequential extraction, and explicit type definitions, mechanisms that help suppress noise during entity and relationship extraction. While the coreference resolution module in CoreKG-no-str-prompt continues to unify repeated or ambiguous legal mentions, its effectiveness is largely overridden by the surge in over-extracted and conflated nodes.

Interestingly, the average noisy node count in CoreKG-no-str-prompt (28.86) is nearly identical to that of GraphRAG (27.43), indicating that removing structured prompts nullifies many of the precision gains achieved through coreference resolution. This can be attributed to the fact that although CoreKG-no-str-prompt extracts fewer nodes overall (68.45 on average) compared to GraphRAG (77.15), it still results in a similar number of noisy extractions (20.15 vs. 21.55). This suggests that while the coreference module helps reduce duplication and ambiguity, the lack of structured guidance leads to attention drift and over-extraction, thereby pushing the overall noise ratio close to the baseline.

In a head-to-head comparison, CoreKG-no-str-prompt shows a consistent rise in noisy nodes across all 20 legal cases compared to CORE-KG, clearly demonstrating the critical role of modified structured prompts in noise reduction within the CORE-KG pipeline.

\subsection{Qualitative Analysis of Extracted Graphs}

Figure~\ref{fig:graph_comparison} presents the generated graphs from CORE-KG, CoreKG-no-coref, and CoreKG-no-str-prompt for Case 6. This section analyzes the structural gains achieved through the incorporation of coreference resolution and structured prompts in the CORE-KG pipeline. To provide a clean visual presentation of the graph, relationship labels are omitted in the visualizations, and node sizes are scaled based on their degree (i.e., the number of relationships connected to that node).

\subsubsection{Analysis of Relationship-to-Node Ratio}

Table~\ref{tab:graph_stats-case06} presents the node count, relationship count, and relationship-to-node ratio (R/N) for graphs generated by CORE-KG, CoreKG-no-coref, and CoreKG-no-str-prompt. To assess the structural quality of these graphs, we compute the R/N ratio, which reflects how densely the nodes are interconnected.

\begin{table}[htbp]
\centering
\caption{Graph Statistics for Case 6}
\label{tab:graph_stats-case06}
\begin{tabular}{lccc}
\toprule
\textbf{Method} & \textbf{\#Nodes} & \textbf{\#Relationships} & \textbf{R/N Ratio} \\
\midrule
CORE-KG      & 32  & 92  & 2.88 \\
CoreKG-no-coref   & 59  & 107 & 1.81 \\
CoreKG-no-str-prompt   & 76  & 127 & 1.67 \\
\bottomrule
\end{tabular}
\end{table}

The graph generated by CoreKG-no-str-prompt contains the largest number of nodes and relationships, with 76 nodes and 127 edges. CoreKG-no-coref produces a moderately sized graph with 59 nodes and 107 edges, while CORE-KG yields a more compact structure containing 32 nodes and 92 edges. CORE-KG achieves the highest R/N ratio of 2.88, indicating a high degree of structural coherence. In contrast, CoreKG-no-coref and CoreKG-no-str-prompt produce lower ratios of 1.81 and 1.67, respectively. Although CoreKG-no-str-prompt produces more nodes, many of them appear loosely connected or isolated, suggesting increased noise. The higher density of the CORE-KG graph demonstrates the combined benefit of coreference resolution and structured prompting in producing a more informative and compact representation.

\subsubsection{Per-Case Graph Comparison and Structural Gains}
We further analyze the graph outputs by visually inspecting the impact of duplicate and noisy node generation. In Figure~\ref{fig:graph_comparison}, duplicate nodes are marked using rectangles, while noisy nodes are enclosed in ovals. Duplicate nodes are colored identically to indicate that they refer to the same real-world entity but appear under different surface forms in the text.

\paragraph{Analysis of CoreKG-no-coref Graph}
Figure~\ref{fig:abl1_graph} shows the graph extracted using CoreKG-no-coref. As presented in Table~\ref{tab:results-duplicate-noise}, CoreKG-no-coref produces a total of 45.85 nodes, which is slightly higher than CORE-KG (41.25) and significantly lower than CoreKG-no-str-prompt (68.45). However, it exhibits a noticeably higher degree of duplication (26.01) compared to CORE-KG (20.27) and CoreKG-no-str-prompt (21.15). This is primarily due to the absence of the coreference resolution module, which fails to unify different surface forms of the same entity and thus causes the model to treat them as distinct nodes.

For example, in the CoreKG-no-coref graph (Figure~\ref{fig:abl1_graph}), the main defendant, “Richard Stone,” appears in several surface variations, such as “Defendant Stone” and “Stone”. Similarly, other person-type entities, including “Eleanor Grant,” “Victor Barnes,” and “Adam Everett”, are each represented as multiple nodes due to these unresolved duplicate mentions.  Additionally, ``John Ed Carlson,'' listed as an ``Undocumented Alien,'' also appears in a shortened form as ``Carlson,'' further contributing to node duplication. This issue is not limited to person-type entities. Route-type entities such as ``Interstate Highway 35'' are inconsistently represented as ``I-35'' and ``Interstate.'' Location-type entities like ``San Antonio'' appear both as ``San Antonio, Texas'' and simply ``San Antonio.'' Likewise, means of transportation such as ``Nissan Maxima'' are scattered across forms like ``Maxima,'' ``Car,'' and ``Vehicle.'' These duplications, among many others, reduce clarity and introduce structural redundancy in the graph, emphasizing the importance of applying coreference resolution prior to knowledge graph construction.

Regarding noise, CoreKG-no-coref performs better than CoreKG-no-struct-prompt and is nearly comparable to CORE-KG. The structured prompt still filters many irrelevant entities and supports more precise extraction. However, several entities remain disconnected from the main graph, as shown in Figure~\ref{fig:abl1_graph}. This occurs for two reasons. 

First, duplicate entities may appear in different surface forms, preventing correct merging. For example, the location “Antonio” is a shortened form of “San Antonio,” but since the merging step relies on matching written names, these are not unified and therefore appear as separate nodes. This observation highlights a limitation in the current entity merging component of the KG construction module, indicating that it could be made more robust.

Second, some nodes lack extracted relationships even when those relationships are present in the text. This occurs due to attention drift, where the model processes many entities in a dense sentence and overlooks certain links, leaving those nodes isolated in the graph. For example, the “65 mile marker” is properly linked to the main graph in CORE-KG (Figure~\ref{fig:corekg_graph}) and CoreKG-no-str-prompt (Figure~\ref{fig:abl2_graph}), but is missing in CoreKG-no-coref (Figure~\ref{fig:abl1_graph}). Structured prompting helps mitigate attention drift. As a result, CoreKG-no-coref contains around 4 isolated nodes, whereas CoreKG-no-str-prompt, which removes the structured prompt, shows approximately 12.

\paragraph{Analysis of CoreKG-no-str-prompt Graph}
Analyzing the CoreKG-no-str-prompt graph, we observe that it performs better than CoreKG-no-coref in terms of duplication. This improvement can be attributed to the presence of the coreference resolution module, which effectively unified surface variations early in the pipeline. However, a few duplicate entities still persist, such as ``Artesia, New Mexico'' and ``Artesia'', ``San Antonio'' and ``San Antonio, Texas'', ``Undocumented Alien'' and ``Undocumented Aliens'', as well as ``Blue Nissan Maxima'', ``Nissan Maxima'', and ``Vehicle''. As discussed in Section~\ref{subsec:rq2corekgnostrprompt}, this duplication is likely due to the use of the standard GraphRAG prompt, which suffers from attention drift and consequently leads to the extraction of irrelevant entities, further contributing to duplication.

However, CoreKG-no-str-prompt exhibits substantially higher noisy entity extraction, since it uses the standard GraphRAG prompt, which lacks the structured instructions that filter out irrelevant or noisy mentions.  As shown in Figure~\ref{fig:abl2_graph}, multiple government-related entities are unnecessarily extracted around nodes like ``Richard Stone'' and ``Victor Barnes.’’ Additionally, several fine-grained but irrelevant entities appear, such as ``Window'' and ``Car Frame,’’ both referring to parts of the ``Nissan Maxima’’. These cases illustrate the importance of in-prompt filtering, sequential extraction, and clearly defined entity types, all key components of CORE-KG’s structured prompt.


One notable observation is that the CoreKG-no-str-prompt graph captures several fine-grained entities that provide valuable context for analyzing the operational nuances of human smuggling networks. For example, entities like ``Passenger Side Backseat,'' where the undocumented alien Carlson was seated, and ``Bushes,'' which concealed Border Patrol Agents along Interstate 35, offer spatial and situational details that enhance the understanding of smuggling tactics. This suggests the need to balance filtering out irrelevant fine-grained entities while retaining those that contribute to deeper analysis.

\paragraph{Analysis of CORE-KG Graph}
Analyzing the CORE-KG graph reveals a well-connected structure with minimal duplication and noise, enabled by the integration of coreference resolution and modified structured prompts. Although some duplicate nodes persist, such as ``Blue Nissan Maxima'' and ``Vehicle'' or ``Undocumented Alien'' and ``Aliens,'' their frequency is significantly lower compared to CoreKG-no-coref and CoreKG-no-str-prompt. Similarly, a few government-related noisy nodes, including ``United States Customs and Border Protection Border Patrol Academy,'' are present but considerably fewer than in the two ablation variants. As a result, CORE-KG produces cleaner and more interpretable graphs that better support downstream analysis.

%

\section{Conclusion}
In this work, we conducted a systematic ablation study of CORE-KG to quantify the individual contributions of its two key components: type-aware coreference resolution and domain-guided structured prompts. By isolating each module, we measured how its removal affects knowledge graph quality in terms of node duplication and legal noise. Removing coreference resolution increases node duplication by 28.25\% and noisy nodes by 4.32\%, while removing structured prompts increases node duplication by 4.29\% and noisy nodes by 73.33\%. These findings demonstrate that both modules are essential and complementary. Structured prompts play a dominant role in improving extraction precision, while coreference resolution enhances entity consolidation by resolving referential variation. This separation in impact highlights the value of modular LLM pipelines, where each component targets a distinct error mode. Future work may explore automating and optimizing prompt construction, evaluate how improved graph quality affects retrieval accuracy and answer coherence in downstream RAG systems, and develop cross-document coreference resolution to construct unified graphs that link recurring actors and patterns across legal cases.





\section*{Acknowledgment}
This material is based upon work supported by the U.S. Department of Homeland Security under Grant Award Number 17STCIN00001-08-00. 
Disclaimer: The views and conclusions contained in this document are those of the authors and should not be interpreted as necessarily representing the official policies, either expressed or implied, of the U.S. Department of Homeland Security.

\bibliographystyle{IEEEtran}
\bibliography{bibliography}
\end{document}